# DIFFERENT ISSUES IN THE DESIGN OF A LEMMATIZER/TAGGER FOR BASQUE

Aduriz I., Alegria I., Arriola J.M., Artola X., Díaz de Ilarraza A., Ezeiza N., Gojenola K., Maritxalar, M.

Department of Computer Languages and Systems - University of the Basque Country

649 p.k., 20080 Donostia - Spain

jipgogak@si.ehu.es

**ABSTRACT**

This paper presents relevant issues that have been considered in the design of a general purpose lemmatizer/tagger for Basque (EUSLEM). The lemmatizer/tagger is conceived as a basic tool necessary for other linguistic applications. It uses the lexical data base and the morphological analyzer previously developed and implemented. Due to the characteristics of the language, the tagset here proposed is structured in four levels so that each level is a refinement of the previous one in the sense that it adds more detailed information. We will focus on the problems found in designing this tagset and on the strategies for morphological disambiguation that will be used.

## 1. Introduction

This paper describes the development of a general purpose lemmatizer/tagger for Basque which will lay the foundations for further applications in the field of automatic processing of Basque texts. In order to elaborate this project the following basic tools will be used:

- The Lexical Database for Basque (LDBB). This database is both source and support for the lexicons needed in different applications, and was designed with the objectives of being neutral in relation to linguistic formalisms, flexible, open and easy to use. At present it contains 60,000 entries, each with its associated linguistic features (category, subcategory, case, number, etc.).

- A general morphological analyzer/generator (Agirre et al., 92). For morphological analysis, we use morphological descriptions based on the two-level formalism proposed by Koskenniemi (Koskenniemi, 83). The morphological analyzer attaches to each input word-form all possible interpretations and its associated information. The LDBB also stores information about the two-level lexicon system.

This morphological analyzer has been extended in two ways so as to make it more robust (Aldezabal et al., 94):

  - The treatment of linguistic variants. Due to the fact that the process of normalization of Basque is still in progress, the morphological processor must deal not only with standard but also with dialectal and non-standard forms of words. For this purpose, new sublexicons and two-level rules dealing with this kind of forms have been added.

  - A two-level mechanism for lemmatization without lexicon, based on the idea used in speech synthesis (Black et al., 91). This mechanism has two main components in order to be capable of treating unknown words: 1) generic lemmas corresponding to each possible open category or subcategory, and 2) two additional rules in order to express the relationship between the generic lemmas at lexical level and any acceptable lemma of Basque.

In section 2 we introduce some special characteristics of Basque morphology and the problem of ambiguity. Section 3 describes the problems found in designing a tagset for Basque texts. After that, in section 4 we will show the strategies that will be used for disambiguation. Finally, in section 5 some conclusions will be presented.

## 2. Brief Description of Basque

Basque is an agglutinative language, that is, for the formation of words the dictionary entry independently takes each of the elements necessary for the different functions (syntactic case included). More specifically, the affixes corresponding to the determinant, number and declension case are taken in this order and independently of each other.

One of the principal characteristics of Basque is its declension system with numerous cases. The marks corresponding to determination, number and case appear only after the last element in the noun phrase. This last element may be the noun, but also typically an adjective or a determiner. For instance, *semeArEN etxeAN* (in the house of the son)

| *seme* | *A* | *r* | *EN* | *etxe* | *AN* |
|---|---|---|---|---|---|
| noun (son) | determinate | epenthetical element | genitive | noun (house) | inessive case |

As prepositional functions are realized by case suffixes inside word-forms, Basque presents a relatively high power to generate inflected word-forms. For instance, from one noun entry a minimum of 135 inflected forms can be generated. Moreover, while 77 of them are simple combinations of number, determination, and case marks, not capable of further inflection, the other 58 are word-forms ended with one of the two possible genitives or with a sequence composed of a case mark and a genitive mark. If the latter is the case, then by adding again the same set of morpheme combinations (135) to each one of those 58 forms a new, complete set of forms could be recursively generated. This kind of construction reveals a noun ellipsis inside a complex noun phrase and could be theoretically extended *ad infinitum*; however, in practice it is not usual to find more than two levels of this kind of recursion in a word-form.

Derivation and composition are quite productive and irregular, being widely used in neologism formation.

### 2.1. Morphological ambiguity in Basque

In order to deal with morphological ambiguity we distinguish the following main types:

- Categorial (or part of speech) ambiguity, like Noun/Verb, Verb/Adjective/Adverb, ... Just to have an idea about the magnitude of this kind of ambiguity, after analyzing a corpus of 2,113 word-forms we found 655 ambiguous, so the rate of categorial ambiguity is around 0.31 (1.38 analysis for each word-form).

- Morphosyntactic ambiguity. There are several possible morphosyntactic interpretations attached to each input word-form. For instance:

    *gizonak*    Nominative Plural / Ergative Singular
    (the men / the man)

    The borderlines between morphology and syntax are not clear, because the information attached to each analysis contains features belonging to both morphology and syntax:

    *ikusiaz*    Instrumental Case at morphological level / Modal Clause at syntactic level
    (seeing)

    Moreover, we must also point out that whenever intraword noun ellipsis occurs, genitive recursion has been applied, but the reverse is not always true. This kind of ambiguity (elliptic ambiguity) will be presented later in section three.

# 3. The design of the tagset

The choice of a tagset is a critical aspect when designing a tagger, because the usefulness of the product and the ambiguity rate will depend on it. Before defining the tagset we have had to take some aspects into account:

a) There was not any exhaustive tagset for automatic use because manual lemmatization processes carried out on Basque texts in previous projects (Urkia and Sagarna, 91) did not include a systematically built tagset.

b) The output of the morphological analyzer, due to the characteristics of the language, is too rich and does not offer a directly applicable tagset.

Let us start by considering the main problems concerned with the second item.

## 3.1. The treatment of the ellipsis

As we have mentioned above, noun ellipsis inside a word-form is a usual phenomenon. When noun ellipsis occurs, there is an explicit lemma plus an implicit one (it is possible to have multiple ellipsis in a single word-form, but it is very infrequent to find more than two). *Semearena* (the one of the son) is an example of a simple ellipsis:

| *seme* | *a* | *r* | *en* | $\varepsilon$[1] | *a* |
|---|---|---|---|---|---|
| noun (son) | determiner | epenthetical element | genitive | (elliptic noun) | nominative case |

If we put it in an actual context:

> *Aitaren etxea handia da eta **semearena** txikia.*
> (The house of the father is big and **the one** (house) **of the son** is small.)

In this example the word *semearena* has an internal ellipsis referring to *house*. Thus, there is an explicit lemma (son) and one implicit lemma (house). There are two choices to tag this word:

- Tagging at word-level. The tag corresponds to the explicit lemma or it may be a composed tag collecting the information of all lemmas (explicit and implicit) present in the word. In this case we will use a composed tag:

    *semearena*/**NOUN_WITH_NOUN_ELLIPSIS**

    *ederrekoetatik*/**ADJECTIVE_WITH_NOUN_ELLIPSIS**
    (from the ones of the pretty one)

- Tagging at lemma-level, so that each explicit or implicit lemma will have its own tag. E.g.:

    *semearen-*/**NOUN**    $\varepsilon$-*a*/**NOUN**

    *ederreko-*/**ADJECTIVE**    $\varepsilon$-*etatik*/**NOUN**

Another important fact is that we can not assert that whenever inflection marks appear after a genitive mark noun ellipsis occurs. In the previous cases *ederrekoetatik* is theoretically ambiguous, although the elliptic interpretation is the most likely. The output of the tagger in the second alternative (lemma-level tagging) is more complex because two possibilities must be reflected:

(1)   *ederrekoetatik*/**ADJECTIVE**     (no ellipsis)
(2)   *ederreko-*/**ADJECTIVE**     $\varepsilon$-etatik/**NOUN**

---

[1]   The symbol $\varepsilon$ stands for the elided noun (implicit lemma).

After taking both choices into account, we took the first option: tagging will be at word-level but the information about the elided components will be kept and expressed by means of composed tags.

## 3.2. Derivation and composition

As we previously mentioned, composition and derivation are very productive. There are two main problems related to them:

- Ambiguity in compounds. As the terms appearing in a compound may be ambiguous, we decided to index the interpretations corresponding to compounds, much in the way done in (Leech et al., 94). This will also have the effect of increasing the ambiguity. For example:

    *plaza*      *gizon*[2]
    /NOUN        /NOUN
                 /VERB
    /NOUN(1)     /NOUN(2)     (indexed tags for compounds)

- Derivation also poses some problems, because many times the derivative morphemes change the category of the stem. In some cases a composed tag will be assigned, whereas in others only the tag corresponding to the resulting category will be kept.

    E.g.:     *etor* /VERB + *tze*/NOMINALIZATION_SUFFIX

              => *etortze*/VERBAL_NOUN[3]     (composed tag)

Both problems have been addressed in the resulting tagset.

## 3.3. The Multilevel Tagset

In designing the general tagset we tried to meet the following requirements:

- It must take into account all the above mentioned problems of ellipsis, derivation and composition.

- In addition, the lemmatizer/tagger has to be general, because it will be used as a step prior to other applications like parsing, information retrieval, indexing, etc.

- It must be coherent with the information provided by the morphological analyzer.

Taking all these considerations into account, the tagset has been structured in four levels, ranging from the simplest part-of-speech tagging scheme up to the full morphological information:

- In the first level, seventeen general categories are included (noun, verb, etc.). This level distinguishes different tags for ellipsis and for the indexing of compounds. It is the basic tagset for ordinary lemmatization.

- In the second level each category tag is further refined by subcategory tags. For example, the verb category has two subcategories: simple and compound verbs.

- The third level may include other interesting morphological information, as declension case, number, etc. In this parametrizable level, the user is allowed to specify which information he/she wishes. This one and the previous level are suitable for the construction of tables of probabilities as well as for the design of rules to be used in disambiguation.

---

[2]  *plaza:*        square
     *gizon:*        man / to become a man
     *plaza gizon:*  public man

[3]  *etortze:*      (the act of) coming

- The output of the morphological analysis constitutes the last level of tagging. The only difference with the previous level is that, here, all the information given by the morphological analyzer is taken into account. This is, obviously, the level that will be the input for syntactic and other kinds of language processing.

## 4. The disambiguation process

In recent years a number of projects have centered on the automatic disambiguation of texts. Our lemmatizer/tagger, as others (Leech et al., 94), will try to combine the two kinds of methods most successfully used:

**Methods based on linguistic knowledge**. The Constraint Grammar formalism (Karlsson et al., 95; Voutilainen, 94; Tapanainen, 94) was designed with the aim of being a language-independent and robust tool to disambiguate and analyze unrestricted texts. It has been used in a project to disambiguate English corpora of about 200 million word-forms, and it claims to have a better performance than statistical methods (about 99% accurate in full disambiguation).

Although the formalism is intended to obtain a surface-syntactic annotation of texts, its main emphasis was put on solving the ambiguity problem. It works on a text where all the possible morphological interpretations have been assigned to each word-form by the morphological analyzer. The role of the CG system is to apply a set of linguistic constraints that discard as many alternatives as possible, leaving in the end fully disambiguated sentences, with one syntactic label for each word-form.

In our case, rules for morphological disambiguation are being devised and tested, focusing on the problem of categorial ambiguity. They will be tested against a previously disambiguated text, before performing a massive disambiguation process.

**Statistical techniques** (DeRose, 88; Garside et al., 87; Cutting et al., 92; Elworthy, 93). The method proposed in different projects uses bigram or trigram probabilities based on knowledge obtained from corpora. Most of the systems use tables of probabilities for each tag of each word-form, but dealing with highly inflected languages implies that the set of word-forms to be considered is wider than in other languages. A possible approach is to build these tables based on the information associated to lemmas instead of word-forms, although the potential loss of information that could arise in this case has to be considered; in this case, the problem is to recognize unambiguously the lemma.

We are dealing with this problem at present. A corpus is being manually disambiguated in order to extract the first tables for automatic disambiguation. The disambiguation is being done at the level of morphological information so that it will be possible to experiment with different tagsets.

In our system, once the morphological analysis has been performed, disambiguation based on linguistic knowledge will help in removing most of the incorrect interpretations. The remaining ambiguity will be dealt with by means of statistical techniques.

## 5. Conclusions

A lemmatizer/tagger for Basque, currently under development, has been described. It is based on a general-purpose morphological analyzer and it uses a mixed disambiguation technique that combines linguistic and statistical knowledge.

In our opinion, syntactic aspects must be considered when tagging words in agglutinative languages, where morphology and syntax are so tied. This characteristic has been taken into account when analyzing different problems —such as intraword ellipsis, derivation, and composition— arising in the design of the system. Particularly, the problem of intraword ellipsis has been described and two different tagging solutions discussed.

Along with the design of the lemmatizer, an incrementally structured four-level tagset that can be adapted to different needs has been defined.

One of the things this work has shown us is that tagsets developed for typologically very different languages are not directly applicable. This is a conclusion that should be taken into account when designing tagset standards.